\documentclass[runningheads,a4paper]{llncs}

\usepackage{amssymb}
\usepackage{graphicx}
\usepackage[caption = false]{subfig}
\usepackage{wrapfig}

\usepackage{url}
\urldef{\mailsa}\path|{yyin, aaa.aaa, bbb.bbb, ccc.ccc,|
\urldef{\mailsb}\path|ddd.ddd, eee.eee}@icadmed.com|    
\newcommand{\keywords}[1]{\par\addvspace\baselineskip
\noindent\keywordname\enspace\ignorespaces#1}

\usepackage{amsmath}

\graphicspath{{./figures/}}

\newif\ifInternalUse
\InternalUsefalse

\begin{document}

\mainmatter  

\title{Transferring Learned Microcalcification Group Detection from 2D Mammography to 3D Digital Breast Tomosynthesis Using a Hierarchical Model and Scope-based Normalization Features}

\titlerunning{Robust Microcalcification Group Detection in Mammography and Tomosynthesis}

%
%
\author{Yin Yin%
\thanks{Sent email to contact author at yinyin.dsp@hotmail.com}%
\and Sergei V. Fotin\and Hrishikesh Haldankar\and Jeffrey W. Hoffmeister\and\\
Senthil Periaswamy}
\authorrunning{Y. Yin, \textit{et al.}}
\titlerunning{Computer-aided detection of microcalcification groups}

\institute{iCAD Inc.,
98 Spit Brook Road, Suite 100, Nashua, New Hampshire, USA
}

%
%

\maketitle

\begin{abstract}
A novel hierarchical model is introduced to solve a general problem of detecting groups of similar objects. Under this model, detection of groups is performed in hierarchically organized layers while each layer represents a scope for target objects. The processing of these layers involves sequential extraction of appearance features for an individual object, consistency measurement features for nearby objects, and finally the distribution features for all objects within the group. Using the concept of scope-based normalization, the extracted features not only enhance local contrast of an individual object, but also provide consistent characterization for all related objects. As an example, a microcalcification group detection system for 2D mammography was developed, and then the learned model was transferred to 3D digital breast tomosynthesis without any retraining or fine-tuning. The detection system demonstrated state-of-the-art performance and detected $96\%$ of cancerous lesions at the rate of $1.2$ false positives per volume as measured on an independent tomosynthesis test set. 
\end{abstract}

\section{Introduction}


In this study, we tackle a commonly seen problem in medical image analysis: how to transfer the knowledge learned from existing training data to a new data without retraining or fine-tuning the system, in the situation when the new data has large intensity variations compared to the training data.  This is especially valuable if the newly acquired data is rare or expensive. For example, deep convolutional neural networks will not be applicable for this task as they require larger amounts of training data.

\begin{figure}[t]
	\centering
	\subfloat[A mammography image]{
		\includegraphics[width=0.45\textwidth,height=4.5cm]{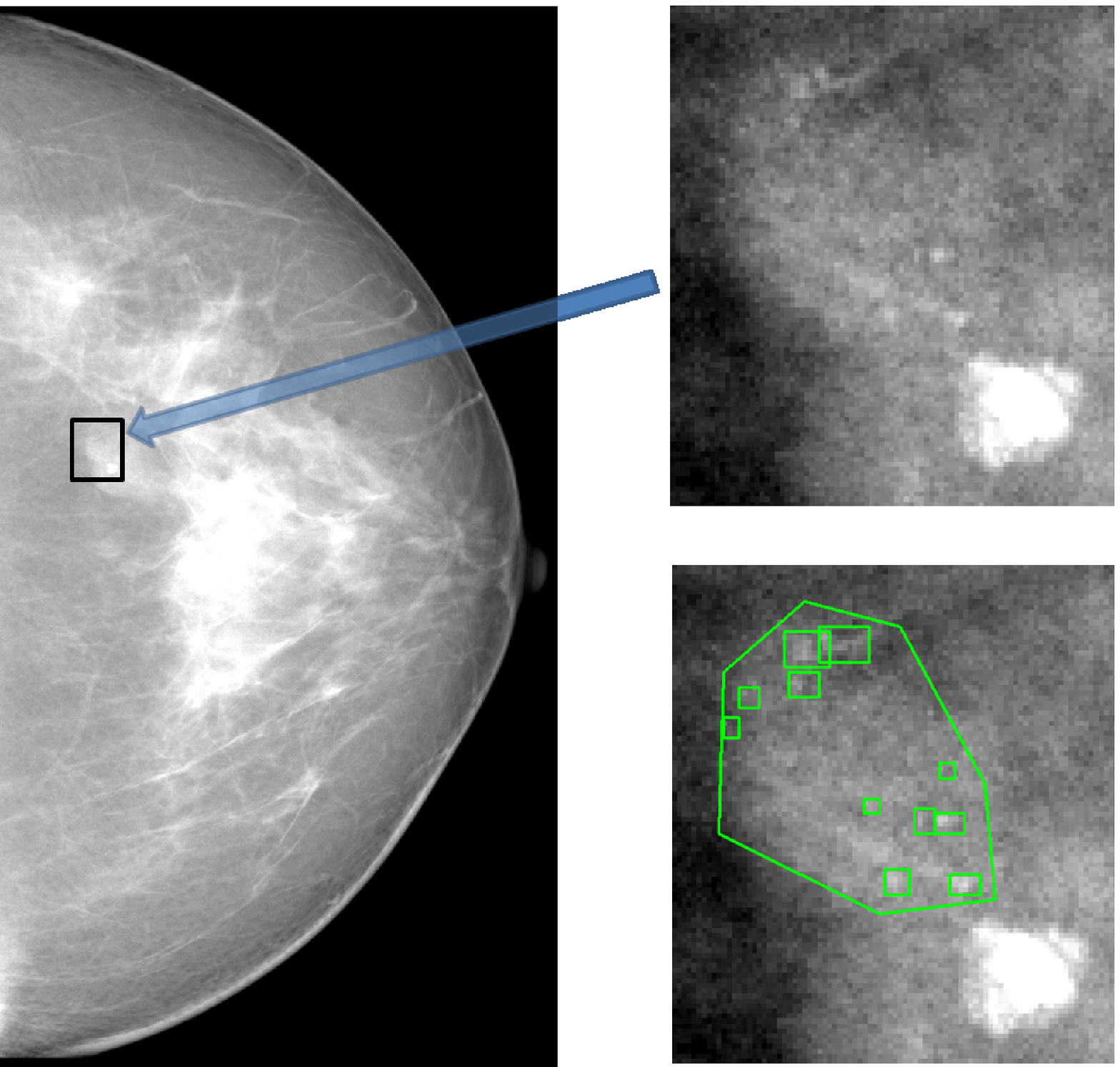}
		}
	\qquad
	\subfloat[A DBT volume]{
		\includegraphics[height=4.5cm]{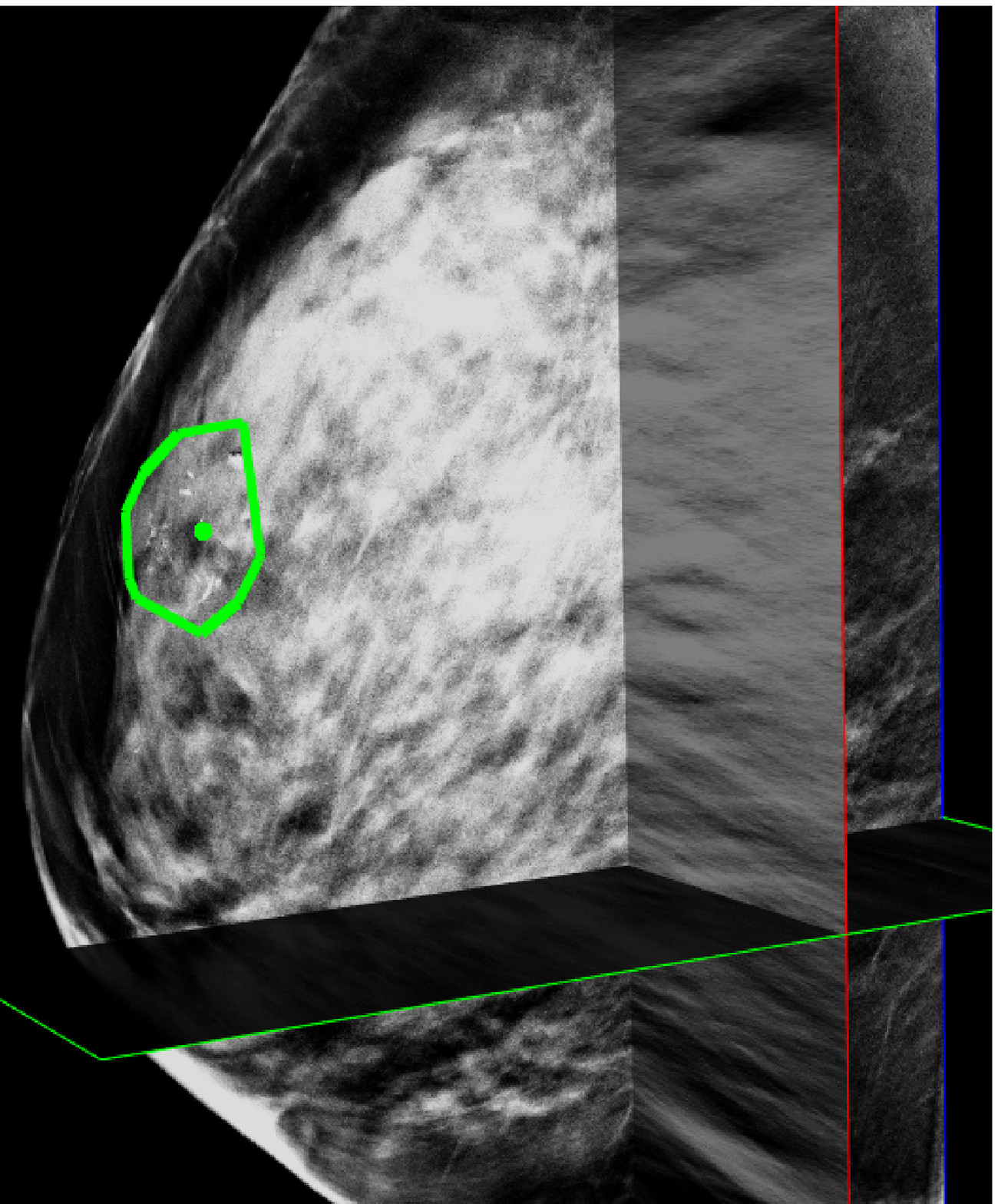}
		}
\caption{(a) Left: An example of a malignant mammography screening image from GE Senographe mammography system showing the difficulty of finding MC groups. Right top: The local MC group region. The region is small and MCs are barely visible even with adjusted window level to enhance contrast. Right bottom: The ground truth of individual MCs (the bounding boxes), and the malignant group (the convex hull contour).   (b) An example of GE SenoClaire DBT volume with manually labeled group contour on one slice (the green contour). The inter-slice boundaries of the group are labeled as green points. 
}
\label{fig:ge_example}
\end{figure}

One application of such a knowledge transfer problem is the detection of microcalcification (MC) groups in 3D digital breast tomosynthesis (DBT). A MC group is composed of multiple small and similar individual MCs, and is considered a possible sign of breast cancer. Because MCs may be small and lack enough contrast (as in Figure~\ref{fig:ge_example}(a)), they can be easily missed by radiologists during the routine screening process.  Needless to say, a MC group computer-aided detection (CAD) can help radiologists locate suspicious regions and help them make diagnostic decisions.

A large number of literature has been reported on creating CAD system for MC groups in 2D mammography (as in Figure~\ref{fig:ge_example}(a)). However, publications on CAD for recently emerged DBT (as in Figure~\ref{fig:ge_example}(b)) are very limited.  This is due to the difficulty of collecting enough cancer cases to train a DBT CAD from scratch.  Instead, most researchers built systems on small DBT datasets with limited training~\cite{BernardDM08,SahinerMP12} or applied a mammography CAD directly on slices or projections of DBT volumes~\cite{ParkSPIE08,ReiserMP08,SamalaSPIE15}.  

Mammography and DBT are different imaging technologies both targeting  abnormalities inside woman breasts. The MCs in mammography images and DBT volumes share similar properties like appearance, shape and spatial distribution in 2D. At the same time they differ greatly in brightness, contrast and presence of another spatial dimension. Building a DBT CAD by transferring common knowledge learned from a large number of easily accessible mammography images can be more reliable than to train a CAD based on a small number of DBTs.  We further argue that the key to perform such knowledge transfer is to build a robust system that overcomes intensity variations within the existing training data.  To achieve this, a multi-layer bottom-up hierarchical model is proposed to solve a general problem of detecting a group of similar objects.   

\section{Methods}
\subsection{The Multi-Layer Hierarchical Model}
\begin{wrapfigure}{L}{0.55\textwidth}
\centering
\includegraphics[width=.5\textwidth, height=.4\textwidth]{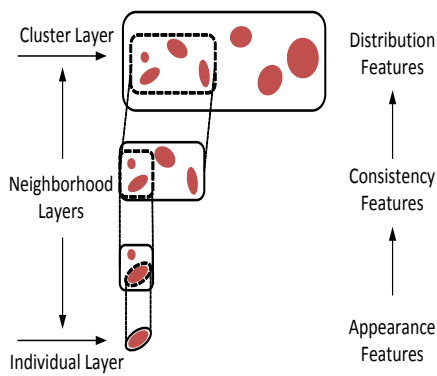}
\caption{The proposed bottom-up hierarchical model.}
\label{fig:model}
\end{wrapfigure}

The proposed hierarchical model is shown in Figure~\ref{fig:model}.  The model is composed of multiple connected layers.  Each layer represents areas of interest associated with the target objects.  The area of interest is the ``scope'' of the object. 


The bottom layer is called ``individual layer'', which focuses on individual objects (or parts of objects).  The scope of this layer is closely around the individual object, and the objects are treated independently from each other.  

The topmost layer is called ``cluster layer'', which represents the entire group of objects. The scope of this layer includes all the objects in the group.

The layers between individual layer and cluster layer are called ``neighborhood layers'' that handle the relationship among objects.  The scopes of these layers depend on the number of objects to compare.  Multiple neighborhood layers can be created with different scopes. 

The model also guides feature extraction at each layer.  Close to the individual layer, appearance features for individual object should be extracted.  Then consistency measurements should be added to the feature set when system processing the neighborhood layers.  Close to the cluster layer, the feature set expands to include distribution measurement of objects within groups. Please note that the model only provides a guidance and there is no clear boundary of where each type of features has to be computed.

Furthermore, the concept of ``scope'' plays an important role to extract robust features within each layer.  One possible usage is to use normalization within each of the local scopes as stated below.

\subsection{Scope-based Normalization}
Image normalization is an effective way to reduce image intensity variations. 
A simple region-based brightness and contrast normalization approach would be to linearly transform all pixel values within the region to have zero mean and unit standard deviation. In contrast to the entire image based normalization that may not be effective to balance object local contrast variations, the scope-based normalization applies to each scope region independently: 

\begin{equation}\label{eq:sn}
 I'(x) = (I(x)-\mu_s)/\sigma_s \quad \forall{x\in{s}}
 \end{equation} 
where $I(x)$ is the input image pixel value inside scope $s$. The mean and standard deviation of all the pixels in $s$ are $\mu_s$ and  $\sigma_s$. $I'$ is the output normalized pixels. Equation \ref{eq:sn} can apply to feature maps and decrease feature value variation, with $I$ being feature values. One recent study showed that normalization of feature maps can also speedup learning process for complicated detection systems~\cite{IoffeARXIV15}.

With scope-based normalization, local contrast of the individual objects is enhanced in the lower level of the layers, and the consistency among neighboring objects can be uniformly evaluated in the neighborhood layers.

\section{MC Group CAD Implementation Details}
Our MC group detection system is built following the guidance of the hierarchical model. The system starts with generating individual MC candidates. After that, four layers are used to remove false positives (FPs): one bottom appearance layer, two neighborhood layers and one top distribution layer.

\ifInternalUse
\else
All features from each of the layers are extracted from pre-computed image feature maps.  The feature maps are: the original input image, the MC enhanced image using grayscale white-top hat operation, the gradient image, and two Laplacian of Gaussian images with different Gaussian kernels.
\fi
   

\ifInternalUse
\else
Many approaches can find bright spots on images as MC candidates with high sensitivity and also with a large number of FPs. Here Frangi's objectness measure~\cite{FrangiMICCAI98} is used to get 2D bright blobs on the grayscale white-top hat image followed by affine shape adaptation~\cite{MikolajczykIJCV04}. The output candidates are described as rotated ellipses, with the blob measure responses, best scales and corresponding eigenvalues.     
\fi

\subsection{Layer Designs}
\subsubsection{Individual Layer}
The scope of the first bottom layer $S_a$ is defined as elliptic regions of generated individual MC candidates. Individual feature maps are first locally normalized within $S_a$ and the regional features are extracted to describe the local appearance of MCs.  These regional features are simply picked as the minimum value, maximum value, skewness and kurtosis from each $S_a$ on each feature map.  Other appearance features include descriptors of the MC shape.  They are candidate's eigenvalues recomputed from normalized image based on scope $S_a$, the ratio of the eigenvalues, the rotation angles, and the scale of the candidates.  The output feature set for this layer is $f_{S_a}$.  

\subsubsection{The First Neighborhood Layer}
The scope of the first neighborhood layer $S_{n1}$ is a $6$-by-$6$~mm bounding box centered at each candidate's center.  This layer is designed as a region containing individual MC and its surrounding context.  For that reason, the scope of this layer is chosen to be larger than the individual layer.  The same appearance features are computed on all the feature maps with normalization based on $S_{n1}$ to form feature set $f_{S_{n1}}$ of this layer. 

\subsubsection{The Second Neighborhood Layer}
Clinically, the grouped MCs are identified within a $1$~cm$^3$ volume. Accordingly, the scope of the second neighborhood layer $S_{n2}$ is selected as regions that contain candidates within $l_{max}=10$~mm range. Specifically, for a candidate $c$, its all neighboring candidates $c'\in{N(c)}$ with $dist(c,c')\leq{l_{max}}$ are found.  $S_{n2}$ is 2D bounding box regions with all $c$ and $c'$s enclosed. Note that $S_{n2}$ may be different from candidate to candidate. 

The feature set $f_{S_{n2}}$ computed in this layer is to measure the consistency of the appearance features $f'_{S_a}$ between $c$ and $c'$s.  $f'_{S_a}$ has the same features as in $f_{S_a}$ but with normalization on scope $S_{n2}$. The difference of the mean and standard deviation between $c$ and $n(c)$ number of candidates in $N(c)$ are measured:
 
\begin{equation*} \label{eq:fn2}
f_{S_{n2}}(c) = \left\{f'_{S_a}(c) - \frac{1}{n(c)}\sum_{c'\in{N(c)}}{f'_{S_a}(c')}, \quad \sqrt{\frac{\sum_{c'\in{N(c)}}(f'_{S_a}(c) - f'_{S_a}(c'))^2}{n(c)}}\right\}.
\end{equation*}

\subsubsection{The Cluster Layer}
The output MC candidates form groups at this layer.  So the scope $S_c$ of this layer is each possible MC group. An agglomerative clustering approach~\cite{RokachHandbook05} is used to form possible groups with a bottom-up fashion following the descending order of dissimilarity value $d$.  Candidates $c_i$ and $c_j$ are expected to be grouped first if they are close in space, have similar appearance, or both are likely from MC groups. Therefore, $d$ is defined as a combination of distance between $c_i$ and $c_j$ and their probabilities $p\in[0,1]$ achieved from the output of the last neighborhood layer: 

\begin{equation*}\label{eq:md}
d(c_i,c_j) =  \frac{dist(c_i, c_j)}{l_{max}} + a|p_{c_i}-p_{c_j}| + b\left(1.0 - \frac{p_{c_i} + p_{c_j}}{2}\right) \quad \text{for $dist(c_i,c_j)\leq{l_{max}}$},
\end{equation*}
where $a$ and $b$ are constants.  $a=b=1$ is chosen, which gives reasonable grouping results during the experiment.

The feature set $f_{S_c}$ for each possible group contains distribution information for the MC candidates inside the group, for example, the eigenvalue of the distribution in space, the area of the group, the density of the candidates, the mean and standard deviation of the dissimilarity measures. $f_{S_c}$ also includes the mean and standard deviation of candidates' appearance features in $f_{S_a}$ normalized based on scope $S_c$. 
 
Once each candidate group gets a classifier score, the final outputs can be generated by picking up the group with the largest probability and removing all other overlapping groups iteratively.

\subsection{Training}
A mammography training dataset with large image intensity variations is collected. The dataset includes images from eleven mammography vendors, and is mixed with traditional film, computed radiography (CR) and digital radiography (DR) mammography images. The total radiologist or mammography expert-truthed biopsy-proven malignant MCs and MC group numbers are $13614$ and $1305$ respectively.  Most of the training images are raw X-ray projections that were passed through a pixel-wise negative log function before inputting to the system.  Some vendors contain a small number of images ready for radiologist screening, which are input for training directly.  Images generated by removing local means from the log-negated projections are also used to increase the variability of the training set.

Because the feature sets $f_{S_a}$ and $f_{S_{n_1}}$ are generated independently, for simplicity, only one classifier is trained on the combined set.  In total, there are three ensemble of randomized decision trees classifiers trained for the feature sets $\{f_{S_a}, f_{S_{n_1}}\}$, $\{f_{S_a}, f_{S_{n_1}}, f_{S_{n_2}}\}$, and $f_{S_c}$. Each classifier corresponds to a threshold to remove FPs.

\subsection{Extension to Detect in 3D DBT Volumes}  
The 3D DBT provides additional depth information.  However, due to the limited angle of X-ray source used in reconstruction, the resolution of the depth is much worse than the resolution within each DBT slice.  

Hence obvious steps are made to adapt 2D mammography solution to 3D DBT without retraining or fine-tuning the system with limited DBT data.  The modifications of the 2D system include making 2D candidate generator work slice-by-slice and change the scopes $S_{n_2}$ and $S_c$ to be in 3D. With the dimension change of the scopes, all the appearance features are now computed in 3D, except the shape and distribution descriptors are still computed in a 2D fashion with the depth information of the candidates ignored. 

\section{Experiments}

\subsection{Performance on Mammography Test Set}
\begin{figure}[t]
	\centering
	\subfloat[Individual MC FROCs]{
		\includegraphics[width=0.5\textwidth,height=5.0cm]{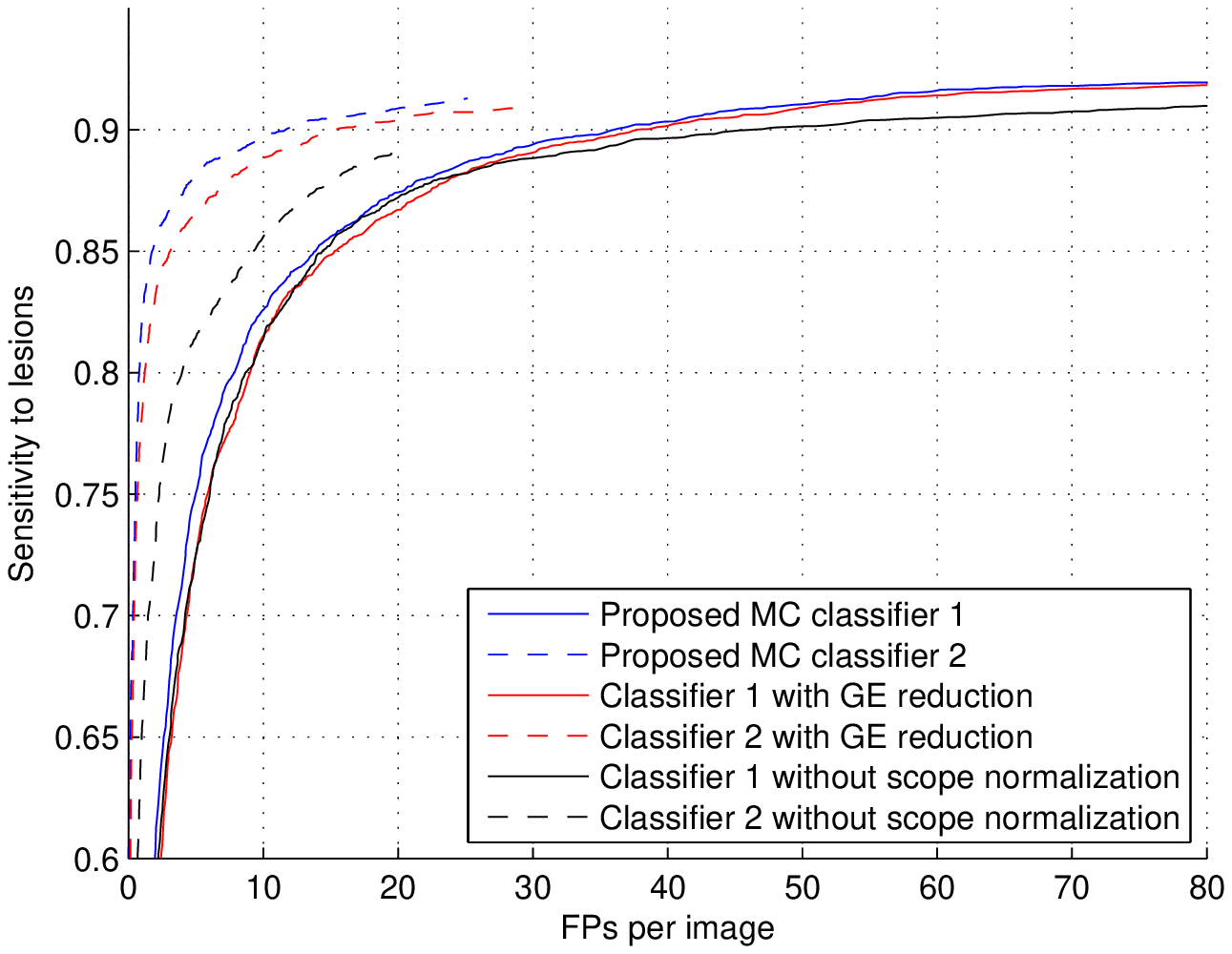}
		}
	\subfloat[MC Group detection FROCs]{
		\includegraphics[width=0.5\textwidth,height=5.0cm]{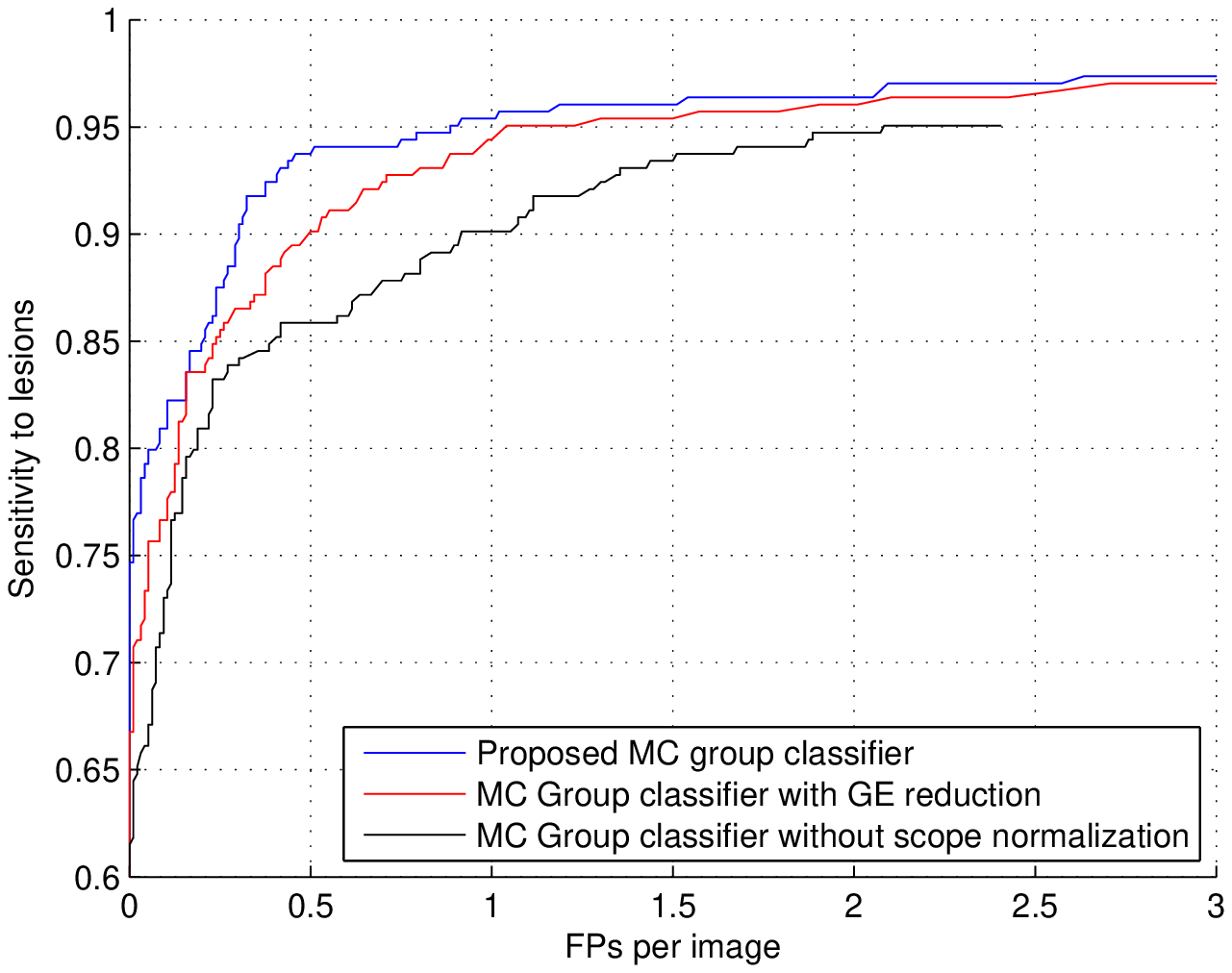}
		}
\caption{(a) The output FROCs from the first individual MC level classifier (trained with feature set $\{f_{S_a}, f_{S_{n_1}}\}$) and the second classifier (trained with feature set $\{f_{S_a}, f_{S_{n_1}}, f_{S_{n_2}}\}$). (b) The output FROCs from the MC group level classifier (trained with feature set $f_{S_c}$). }
\label{fig:ge_test_froc}
\end{figure}

The proposed CAD were first evaluated on a separate GE Senographe projection image dataset with $5683$ and $304$ truthed biopsy-proven malignant MCs and MC groups. The candidate generator provided $92\%$ sensitivity to individual MCs with $1500$ FPs per image. Figure~\ref{fig:ge_test_froc} shows the free-response receiver operating characteristic (FROC) curves of the test set from the outputs of the three pre-trained classifiers.  

To test the system robustness with respect to unseen data, all GE projection images and their counterparts with subtracted local means were removed from the training set.  This reduced about $33\%$ of all the positive training samples. The FROC in red curve in Figure~\ref{fig:ge_test_froc} shows that the system still keeps a high sensitivity rate even with this smaller training dataset.

The black FROC curves were generated from training features without scope-based normalization. Instead, only a global entire breast region based normalization was used for each image. Especially, the FROCs in Figure~\ref{fig:ge_test_froc}(a) show the strength of the scope-based normalization to increase the robustness of both appearance features and consistency measurement features. 
  
\subsection{Performance on DBT Volume Image}
\begin{figure}[t]
	\centering
	\subfloat[The DBT test set MC group FROC]{
		\includegraphics[width=0.5\textwidth,height=4.5cm]{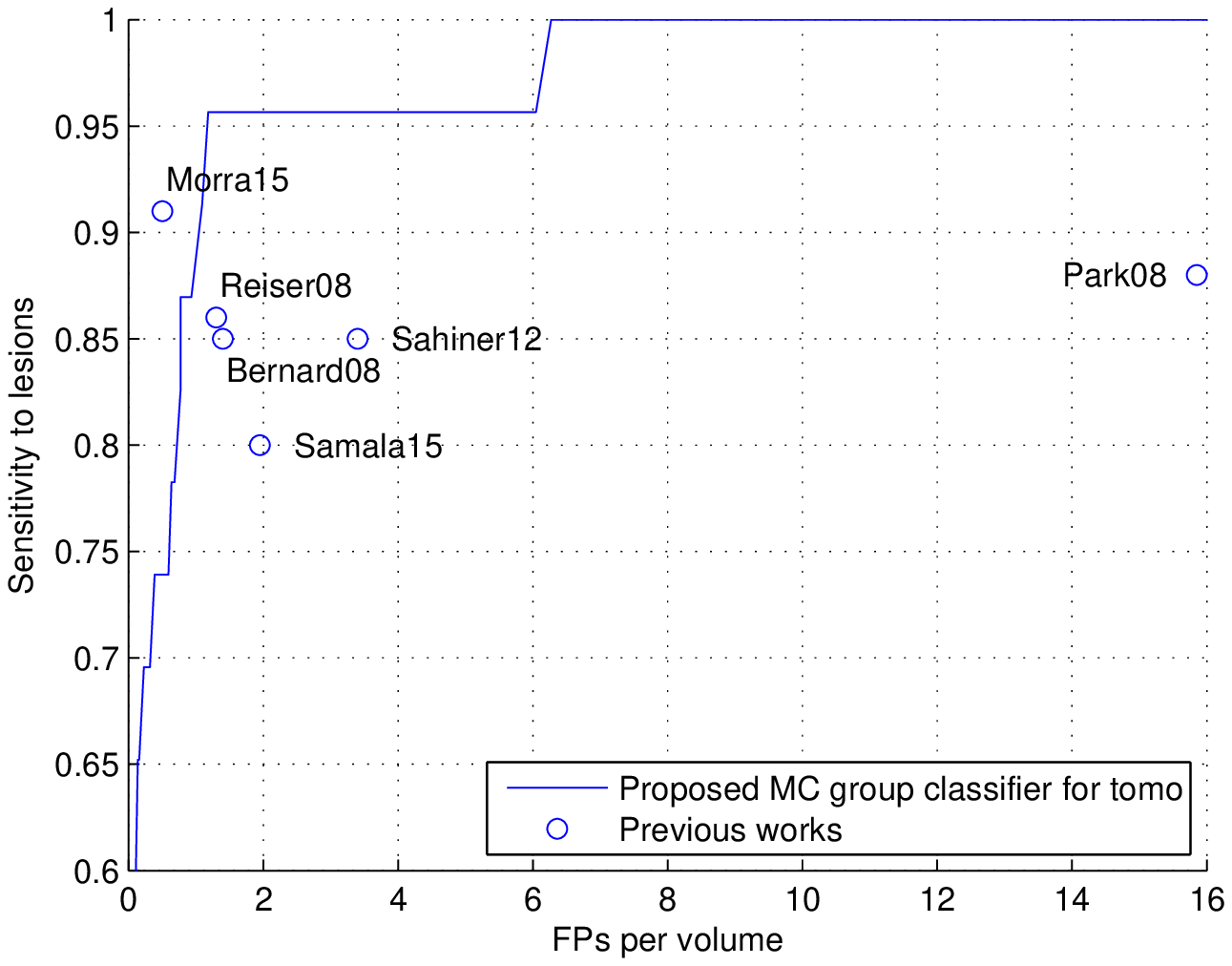}
		}
		\qquad
	\subfloat[Detection examples]{
		\includegraphics[width=0.25\textwidth,height=4.5cm]{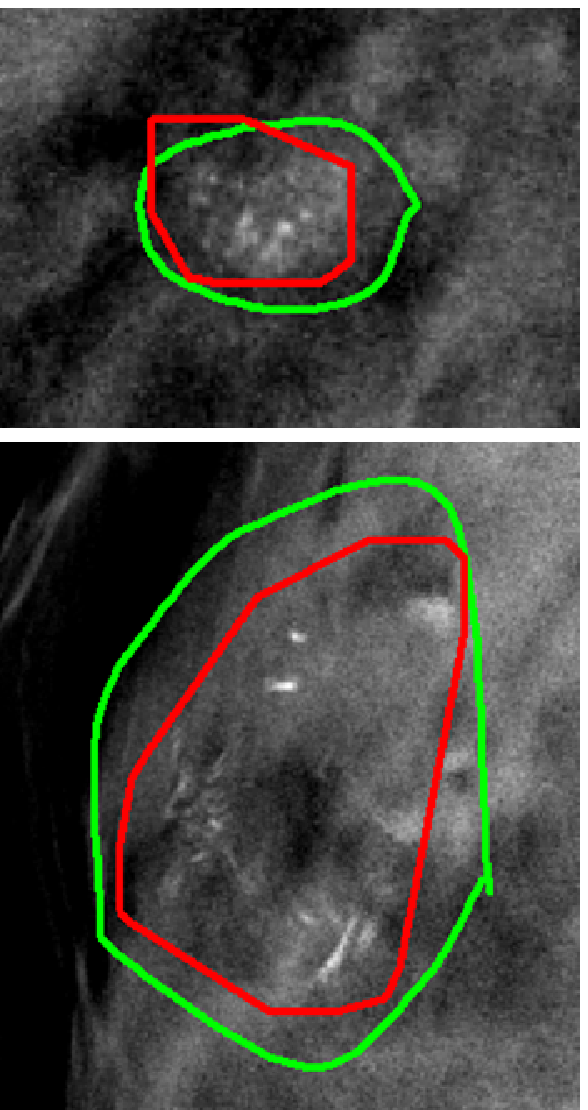}
		}
\caption{(a) Malignant MC group FROC tested on independent DBT volume set with comparison to other reported performances. (b) Examples of the output detections (red) and their corresponding truth contours (green) displayed in DBT slices.}

\label{fig:ge_tomo_performance}
\end{figure}

The proposed system was tested on an independent GE SenoClaire DBT volume set with 23 labeled malignant MC groups and 42 normal cases, no individual MC labeled and no other abnormalities visible. Figure~\ref{fig:ge_tomo_performance}(a) shows the MC group detection FROC and the comparison with other previously published MC group detection performances on DBT~\cite{BernardDM08,SahinerMP12,ParkSPIE08,ReiserMP08,SamalaSPIE15,MorraR15}. Figure~\ref{fig:ge_tomo_performance}(b) shows two successful detections.  The proposed system shows state-of-the-art performance compared with these published works, although all these works including ours were tested on different and relatively small datasets ($\leq23$ cancerous lesions). We need to emphasize that there was absolutely no training or fine-tuning on the DBT data for our system, while other works were not restricted to develop on small numbers of DBTs that could cause overfitting.  The study from Morra \textit{et al.}~\cite{MorraR15} reported better performance, but the authors used Hologic cases with two-view DBT volumes per breast, which may double the chance to find MC groups, while the GE set we were using had only one DBT view available. Furthermore, there is no method details disclosed in~\cite{MorraR15}.

\section{Conclusion}
A CAD for detection of MC groups was built based on 2D mammography images under the guidance of a novel hierarchical model, and features generated by scope-based normalization were used.  The experiment shows the developed CAD system is robust to intensity variations across different images.  Based on that, this CAD can be easily extended to detect MC groups in 3D DBT volumes without acquiring additional DBT training data, and still shows state-of-the-art performance. Further studies of using this model to test on larger dataset or to distinguish between malignant and benign MC groups are needed.


\begin{thebibliography}{4}





\bibitem{BernardDM08} Bernard, S., et al.:Computer-aided microcalcification detection on digital breast tomosynthesis data: A preliminary evaluation. Digital Mammography. 151--157 (2008)

\bibitem{SahinerMP12} Sahiner B., Chan HP., Hadjiiski L., Helvie M., Wei J., Zhou C. and Lu Y.: Computer-aided detection of clustered microcalcifications in digital breast tomosynthesis: a 3D approach. Medical Physics. 39(1), 28--39 (2012)


\bibitem{ParkSPIE08} Park, S. C., Zheng, B., Wang, X.-H. and Gur, D.: Applying a 2D based CAD scheme for detecting micro-calcification clusters using digital breast tomosynthesis images: An assessment. SPIE Med. Imaging Proc. 6915, 691507 (2008)


\bibitem{ReiserMP08} Reiser, I., et al.: Automated detection of microcalcification clusters for digital breast tomosynthesis using projection data only: a preliminary study. Medical Physics. 35(4), 1486--93 (2008) 

\bibitem{SamalaSPIE15} Samala, R. K., Chan, HP., Lu, Y., Hadjiiski, L., Wei, J. and Helvie, M.: Digital breast tomosynthesis: application of 2D digital mammography CAD to detection of microcalcification clusters on planar projection image. SPIE. 9414, 941418 (2015)


\bibitem{IoffeARXIV15} Ioffe S. and Szegedy C.: Batch normalization: accelerating deep network training by reducing internal covariate shift. arXiv:1502.03167v3 (2015)


\bibitem{FrangiMICCAI98} Frangi, A. F., Niessen, W. J., Vincken, K. L. and Viergever, M. A.:Multiscale vessel enhancement filtering. MICCAI. 130--137 (1998) 

\bibitem{MikolajczykIJCV04} Mikolajczyk, K. and Schmid, C.:Scale and affine invariant interest point detectors. International Journal of Computer Vision (IJCV). 60(1),63--86 (2004)


\bibitem{RokachHandbook05} Rokach, L. and Maimon, O.: "Clustering methods." Data mining and knowledge discovery handbook. Springer US. 321--352 (2005) 







\bibitem{MorraR15} Morra L., et al.: Breast cancer: computer-aided detection with digital breast tomosynthesis. Radiology. 277(1), 56--63 (2015)

\end{thebibliography}
\end{document}